\documentclass[runningheads]{llncs}
\usepackage{graphicx}
\usepackage{amsmath,amssymb} 
\usepackage{color}
\usepackage[width=122mm,left=12mm,paperwidth=146mm,height=193mm,top=12mm,paperheight=217mm]{geometry}
\begin{document}
\pagestyle{headings}
\mainmatter
\def\ECCV18SubNumber{1504}  

\title{Disentangling Factors of Variation with Cycle-Consistent Variational Auto-Encoders} 

\titlerunning{Disentangling Factors of Variation with Cycle-Consistent VAEs}

\authorrunning{Jha \textit{et al.}}

\author{Ananya Harsh Jha\inst{1} \and Saket Anand\inst{1} \and Maneesh Singh\inst{2} \and VSR Veeravasarapu\inst{2}\\ \email{\{ananyaharsh12018, anands\}@iiitd.ac.in},\\ \email{maneesh.singh@verisk.com}, \email{vsr.veera@gmail.com}}
\institute{IIIT Delhi \and Verisk Analytics}

\maketitle

\begin{abstract}
Generative models that learn disentangled representations for different factors of variation in an image can be very useful for targeted data augmentation. By sampling from the disentangled latent subspace of interest, we can efficiently generate new data necessary for a particular task. Learning disentangled representations is a challenging problem, especially when certain factors of variation are difficult to label. In this paper, we introduce a novel architecture that disentangles the latent space into two complementary subspaces by using only weak supervision in form of pairwise similarity labels. Inspired by the recent success of cycle-consistent adversarial architectures, we use cycle-consistency in a variational auto-encoder framework. Our non-adversarial approach is in contrast with the recent works that combine adversarial training with auto-encoders to disentangle representations. We show compelling results of disentangled latent subspaces on three datasets and compare with recent works that leverage adversarial training. 

\end{abstract}

\section{Introduction}\label{sec:intro}

Natural images can be thought of as samples from an unknown distribution conditioned on different factors of variation. The appearance of objects in an image is influenced by these factors that may correspond to shape, geometric attributes, illumination, texture and pose. Based on the task at hand, like image classification, many of these factors serve as a distraction for the prediction model and are often referred to as nuisance variables. One way to mitigate the confusion caused by uninformative factors of variation is to design representations that ignore all nuisance variables \cite{DBLP:journals/corr/EdwardsS15,DBLP:journals/corr/LouizosSLWZ15}. This approach, however, is limited by the quantity and quality of training data available. Another way is to train a classifier to learn representations, invariant to uninformative factors of variation, by providing sufficient diversity via data augmentation \cite{Krizhevsky:2012:ICD:2999134.2999257}. 

Generative models that are driven by a disentangled latent space can be an efficient way of controlled data augmentation. Although Generative Adversarial Networks (GANs) \cite{DBLP:journals/corr/GoodfellowPMXWOCB14,DBLP:journals/corr/RadfordMC15} have proven to be excellent at generating new data samples, vanilla GAN architecture does not support inference over latent variables. This prevents control over different factors of variation during data generation. DNA-GANs \cite{DBLP:journals/corr/abs-1711-05415} introduce a fully supervised architecture to disentangle factors of variation, however, acquiring labels for each factor, even when possible, is cumbersome and time consuming.

Recent works \cite{DBLP:journals/corr/MathieuZSRL16,DBLP:journals/corr/abs-1711-02245} combine auto-encoders with adversarial training to disentangle informative and uninformative factors of variation and map them onto separate sets of latent variables. The informative factors, typically specified by the task of interest, are associated with the available source of supervision, e.g. class identity or pose, and are referred to as the \emph{specified} factors of variation. The remaining uninformative factors are grouped together as \emph{unspecified} factors of variation. Learning such a model has two benefits: first, the encoder learns to factor out nuisance variables for the task under consideration, and second, the decoder can be used as a generative model that can generate novel samples with controlled specified and randomized unspecified factors of variation.

\begin{figure}
\centering
\includegraphics[height=5.0cm]{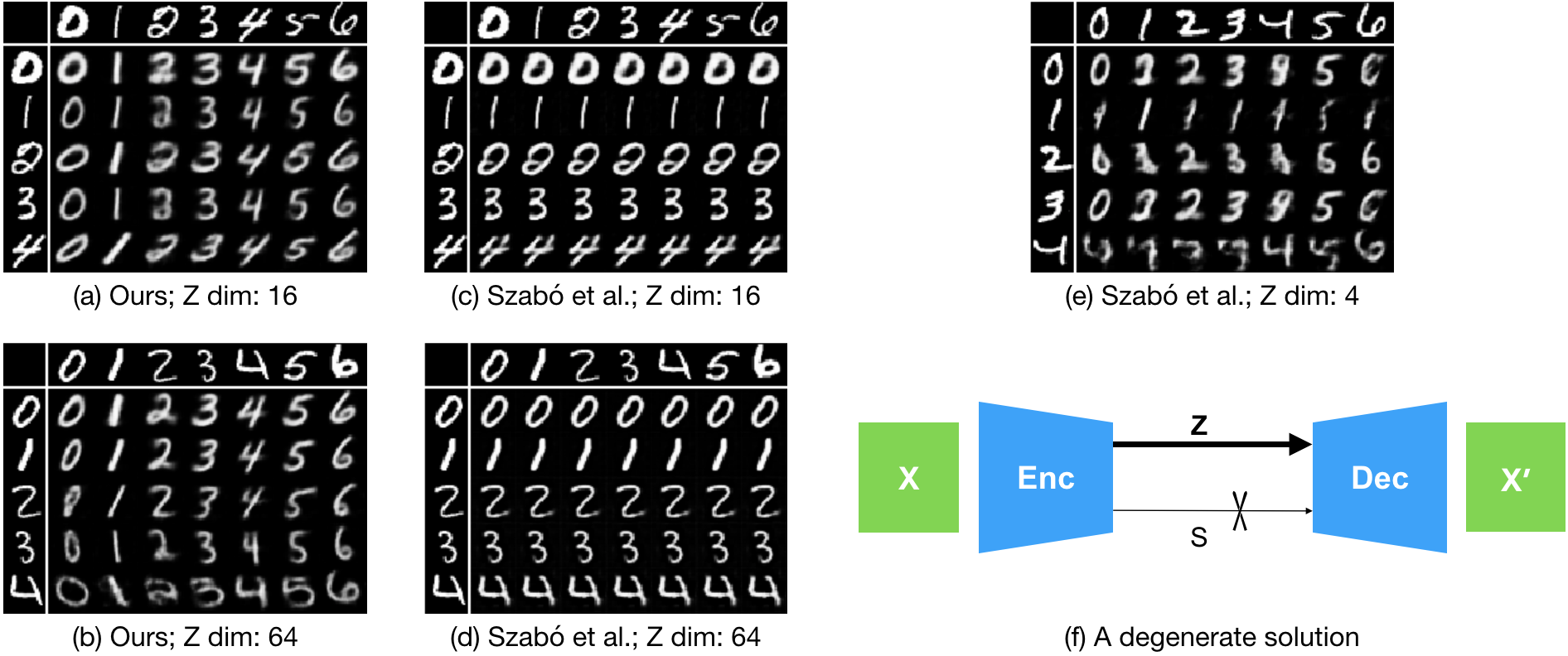}
\caption{\textbf{$s$: specified factors space (class identity), $z$: unspecified factors space.} In each of the image grids: (a), (b), (c), (d) and (e), the digits in the top row and the first column are taken from the test set. Digits within each grid are generated by taking $s$ from the top row and $z$ from the first column. (a) and (b): results of disentangling factors of variation using our method. (c) and (d): results of the non-adversarial architecture from \cite{DBLP:journals/corr/abs-1711-02245}. (e): dimensionality of $z$ required to produce even a few plausible digits using the non-adversarial approach in \cite{DBLP:journals/corr/abs-1711-02245}. (f): visualization of a degenerate solution in case of auto-encoders.}
\label{fig:degenerate_ex}
\end{figure}

In context of disentangled latent representations, Mathieu et al. \cite{DBLP:journals/corr/MathieuZSRL16} define \emph{degenerate solution} as a failure case, where the specified latent variables are entirely ignored by the decoder and all information (including image identity) is taken from the unspecified latent variables during image generation (Fig. \ref{fig:degenerate_ex} (c) and (d)). This degeneracy is expected in auto-encoders unless the latent space is somehow constrained to preserve information about the specified and unspecified factors in the corresponding subspaces. Both \cite{DBLP:journals/corr/MathieuZSRL16} and \cite{DBLP:journals/corr/abs-1711-02245} circumvent this issue by using an adversarial loss that trains their auto-encoder to produce images whose identity is defined by the specified latent variables instead of the unspecified latent variables. While this strategy produces good quality novel images, it may train the decoder to \emph{ignore any leakage of information} across the specified and unspecified latent spaces, rather than training the encoder to restrict this leakage.

Szab{\'{o}} et al. \cite{DBLP:journals/corr/abs-1711-02245} have also explored a non-adversarial approach to disentangle factors of variation. They demonstrate that severely restricting the dimensionality of the unspecified latent space discourages the encoder from encoding information related to the specified factors of variation in it. However, the results of this architecture are extremely sensitive to the dimensionality of the unspecified space. As shown in Fig. \ref{fig:degenerate_ex} (e), even slightly plausible results require careful selection of dimensionality.

\noindent Based on these observations, we make the following contributions in this work:

\begin{itemize}
\item We introduce \emph{cycle-consistent variational auto-encoders}, a weakly supervised generative model, that disentangles specified and unspecified factors of variation using only pairwise similarity labels
\item We empirically show that our proposed architecture avoids \emph{degeneracy} and is robust to the choices of dimensionality of both the specified and unspecified latent subspaces
\item We claim and empirically verify that cycle-consistent VAEs produce highly disentangled latent representations by explicitly training the encoder to reduce leakage of specified factors of variation into the unspecified subspace 
\end{itemize}

To our knowledge, cycle-consistency has neither been applied to the problem of disentangling factors of variation nor has been used in combination with variational auto-encoders. The remaining paper is organized as follows: Sec. \ref{sec:related_works} discusses the previous works relevant in context of this paper, Sec. \ref{sec:cycle_vae} provides the details of our proposed architecture, Sec. \ref{sec:experiments} empirically verifies each of our claims using quantitative and qualitative experiments, and Sec. \ref{sec:conclusion} concludes this paper by summarizing our work and providing a scope for further development of the ideas presented.

\section{Related Work}\label{sec:related_works}

\textbf{Variational Auto-Encoders.} Kingma et al. \cite{DBLP:journals/corr/KingmaW13} present a variational inference approach for an auto-encoder based latent factor model. Let $X = \{x_i\}_{i=1}^N$ be a dataset containing N i.i.d samples, each associated with a continuous latent variable $z_i$ drawn from some prior $p(z)$, usually having a simple parametric form. The approximate posterior $q_{\phi}(z|x)$ is parameterized using the encoder, while the likelihood term $p_\theta(x|z)$ is parameterized by the decoder. The architecture, popularly known as Variational Auto-Encoders (VAEs), optimizes the following variational lower-bound:

\begin{equation} \label{eq:1}
\mathcal{L}(\theta, \phi; x) = \mathbb{E}_{q_{\phi}(z|x)}[\text{log
} \hspace{0.1cm} p_\theta(x|z)] - \text{KL}(q_{\phi}(z|x) \hspace{0.1cm} \| \hspace{0.1cm} p(z))
\end{equation}

The first term in the RHS is the expected value of the data likelihood, while the second term, the KL divergence, acts as a regularizer for the encoder to align the approximate posterior with the prior distribution of the latent variables. By employing a clever linear transformation based reparameterization, the authors enable end-to-end training of the VAE using back-propagation. At test time, VAEs can be used as a generative model by sampling from the prior $p(z)$ followed by a forward pass through the decoder. Our architecture uses the VAE framework to model the unspecified latent subspace.

\noindent \textbf{Generative Adversarial Networks.} GANs \cite{DBLP:journals/corr/GoodfellowPMXWOCB14} have been shown to model complex, high dimensional data distributions and generate novel samples from it. They comprise of two neural networks, a generator and a discriminator, that are trained together in a min-max game setting, by optimizing the loss in Eq. (\ref{eq:2}). The discriminator outputs the probability that a given sample belongs to true data distribution as opposed to being a sample from the generator. The generator tries to map random samples from a simple parametric prior distribution in the latent space to samples from the true distribution. The generator is said to be successfully trained when the output of the discriminator is $\frac{1}{2}$ for all generated samples. DCGANs \cite{DBLP:journals/corr/RadfordMC15} use CNNs to replicate complex image distributions and are an excellent example of the success of adversarial training. 

\begin{equation} \label{eq:2}
\underset{G}{\min} \hspace{0.1cm} \underset{D}{\max} \hspace{0.1cm} V(D, \hspace{0.1cm} G) = \mathbb{E}_{x \sim p_{data} (x)} [\text{log} \hspace{0.1cm} D(x)] + \mathbb{E}_{z \sim p_z (z)} [\text{log} \hspace{0.1cm} (1 - D(G(z)))]
\end{equation}

Despite their ability to generate high quality samples when successfully trained, GANs require carefully designed tricks to stabilize training and avoid issues like mode collapse. We do not use adversarial training in our proposed approach, however, recent works of Mathieu et al. \cite{DBLP:journals/corr/MathieuZSRL16} and Szab{\'{o}} et al. \cite{DBLP:journals/corr/abs-1711-02245} have shown interesting application of adversarial training for disentangling latent factors. 

\noindent \textbf{Cycle-Consistency.} Cycle-consistency has been used to enable a Neural Machine Translation system to learn from unlabeled data by following a closed loop of machine translation \cite{DBLP:conf/nips/HeXQWYLM16}. Zhou et al. \cite{DBLP:journals/corr/ZhouKAHE16} use cycle-consistency to establish cross-instance correspondences between pairs of images depicting objects of the same category. Cycle-consistent architectures further find applications in depth estimation \cite{monodepth17}, unpaired image-to-image translation \cite{DBLP:journals/corr/ZhuPIE17} and unsupervised domain adaptation \cite{DBLP:journals/corr/abs-1711-03213}. We leverage the idea of cycle-consistency in the unspecified latent space and explicitly train the encoder to reduce leakage of information associated with specified factors of variation.

\noindent \textbf{Disentangling Factors of Variation.} Initial works like \cite{Ghahramani:1994:FLE:2998687.2998764} utilize the E-M framework to discover independent factors of variation which describe the observed data. Tenenbaum et al. \cite{Tenenbaum:2000:SSC:1121517.1121518} learn bilinear maps from style and content parameters to images. More recently, \cite{DBLP:journals/corr/abs-1210-5474,DBLP:conf/icml/ReedSZL14,DBLP:conf/icml/TangSH12a} use Restricted Boltzmann Machines to separately map factors of variation in images. Kulkarni et al. \cite{Kulkarni:2015:DCI:2969442.2969523} model \emph{vision as inverse graphics} problem by proposing a network that disentangles transformation and lighting variations. In \cite{DBLP:journals/corr/EdwardsS15} and \cite{DBLP:journals/corr/LouizosSLWZ15}, invariant representations are learnt by factoring out the nuisance variables for a given task at hand.

Tran et al. \cite{DBLP:journals/corr/TranYL17} utilize identity and pose labels to disentangle facial identity from pose by using a modified GAN architecture. SD-GANs \cite{DBLP:journals/corr/DonahueBML17} introduce a siamese network architecture over DC-GANs \cite{DBLP:journals/corr/RadfordMC15} and BE-GANs \cite{DBLP:journals/corr/BerthelotSM17}, that simultaneously generates pairs of images with a common identity but different unspecified factors of variation. However, like vanilla GANs they lack any method for inference over the latent variables. Reed et al. \cite{DBLP:conf/nips/ReedZZL15} develop a novel architecture for visual analogy making, which transforms a query image according to the relationship between the images of an example pair.

DNA-GANs \cite{DBLP:journals/corr/abs-1711-05415} present a fully supervised approach to learn disentangled representations. Adversarial auto-encoders \cite{44904} use a semi-supervised approach to disentangle style and class representations, however, unlike the methods of \cite{DBLP:journals/corr/MathieuZSRL16}, \cite{DBLP:journals/corr/abs-1711-02245} and ours, they cannot generalize to unseen object identities. Hu et al. \cite{DBLP:journals/corr/abs-1711-07410} present an interesting approach that combines auto-encoders with adversarial training to disentangle factors of variation in a fully unsupervised manner. However, the quality of disentanglement still falls short in comparison to \cite{DBLP:journals/corr/MathieuZSRL16,DBLP:journals/corr/abs-1711-02245}.  

Our work builds upon the network architectures introduced by Mathieu et al. \cite{DBLP:journals/corr/MathieuZSRL16} and Szab{\'{o}} et al. \cite{DBLP:journals/corr/abs-1711-02245}. Both of them combine auto-encoders with adversarial training to disentangle specified and unspecified factors of variation based on a single source of supervision, like class labels. Our work differs from these two by introducing a non-adversarial approach to disentangle factors of variation under a weaker source of supervision which uses only pairwise similarity labels.

\section{Cycle-Consistent Variational Auto-Encoders}\label{sec:cycle_vae}

In this section, we describe our model architecture, explain all its components and develop its training strategy.

\subsection{Cycle-Consistency}

\begin{figure}
\centering
\includegraphics[height=2.2cm]{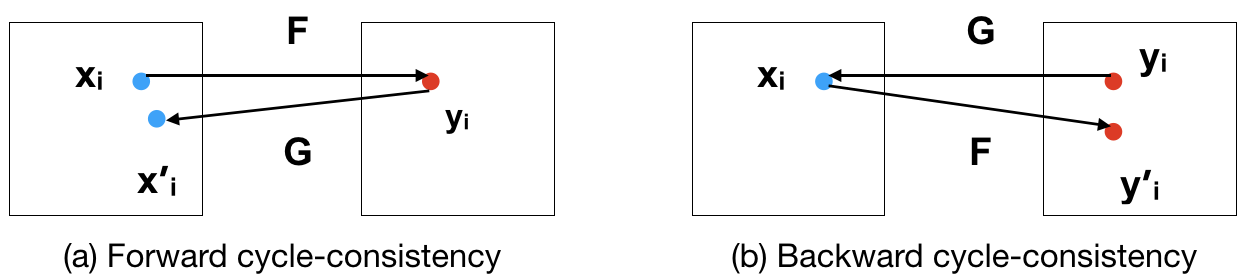}
\caption{(a): Forward cycle in a cycle-consistent framework: $x_i \rightarrow F(x_i) \rightarrow G(F(x_i)) \rightarrow x_i'$. (b): Backward cycle in a cycle-consistent framework: $y_i \rightarrow G(y_i) \rightarrow F(G(y_i)) \rightarrow y_i?'$.}
\label{fig:cycle_consistent}
\end{figure}

The intuition behind a cycle-consistent framework is simple -- the forward and reverse transformations composited together in any order should approximate an identity function. For the forward cycle, this translates to a forward transform $F(x_i)$ followed by a reverse transform $G(F(x_i)) = x_i'$, such that $x_i' \simeq x_i$. The reverse cycle should ensure that a reverse transform followed by a forward transform yields $F(G(y_i))=y_i'\simeq y_i$. The mappings $F(\cdot)$ and $G(\cdot)$ can be implemented using neural networks with training done by minimizing the $\ell_p$ norm based \emph{cyclic} loss defined in Eq. (\ref{eq:3}). 

Cycle-consistency naturally fits into the (variational) auto-encoder training framework, where the KL divergence regularized reconstruction comprises the $\mathcal{L}_{forward}$. We also use the reverse cycle-consistency loss to train the encoder to disentangle better. As is typical for such loss functions, we train our model by alternating between the forward and reverse losses. We discuss the details in the sections that follow.

\begin{equation} \label{eq:3}
\begin{split}
	\mathcal{L}_{cyclic} = \mathcal{L}_{forward}\hspace{0.1cm} &+\hspace{0.1cm} \mathcal{L}_{reverse}\\
	\mathcal{L}_{cyclic} = \mathbb{E}_{x \sim p(x)}[|| \hspace{0.1cm} G(F(x)) - x \hspace{0.1cm} ||_{p}] &+ \mathbb{E}_{y \sim p(y)}[|| \hspace{0.1cm} F(G(y)) - y \hspace{0.1cm} ||_{p}]
\end{split}
\end{equation}

\subsection{Model Description}

We propose a conditional variational auto-encoder based model, where the latent space is partitioned into two \emph{complementary} subspaces: $s$, which controls specified factors of variation associated with the available supervision in the dataset, and $z$, which models the remaining unspecified factors of variation. Similar to Mathieu et al.'s \cite{DBLP:journals/corr/MathieuZSRL16} work we keep $s$ as a real valued vector space and $z$ is assumed to have a standard normal prior distribution $p(z) = \mathcal{N}(0, I)$. Such an architecture enables explicit control in the specified subspace, while permitting random sampling from the unspecified subspace. We assume marginal independence between $z$ and $s$, which implies complete disentanglement between the factors of variation associated with the two latent subspaces. 

\noindent \textbf{Encoder.} The encoder can be written as a mapping $Enc(x) = (f_z(x), \hspace{0.1cm} f_s(x))$, where $f_z(x) = (\mu,\hspace{0.1cm} \sigma) = z$ and $f_s(x) = s$. Function $f_s(x)$ is a standard encoder with real valued vector latent space and $f_z(x)$ is an encoder whose vector outputs parameterize the approximate posterior $q_{\phi}(z|x)$. Since the same set of features extracted from $x$ be used to create mappings to $z$ and $s$, we define a single encoder with shared weights for all but the last layer, which branches out to give outputs of the two functions $f_z(x)$ and $f_s(x)$. 

\noindent \textbf{Decoder.} The decoder, $x' = Dec(z, \hspace{0.1cm} s)$, in this VAE is represented by the conditional likelihood $p_\theta(x|z, s)$. Maximizing the expectation of this likelihood w.r.t the approximate posterior and $s$ is equivalent to minimizing the squared reconstruction error.

\begin{figure}
\centering
\includegraphics[height=3.5cm]{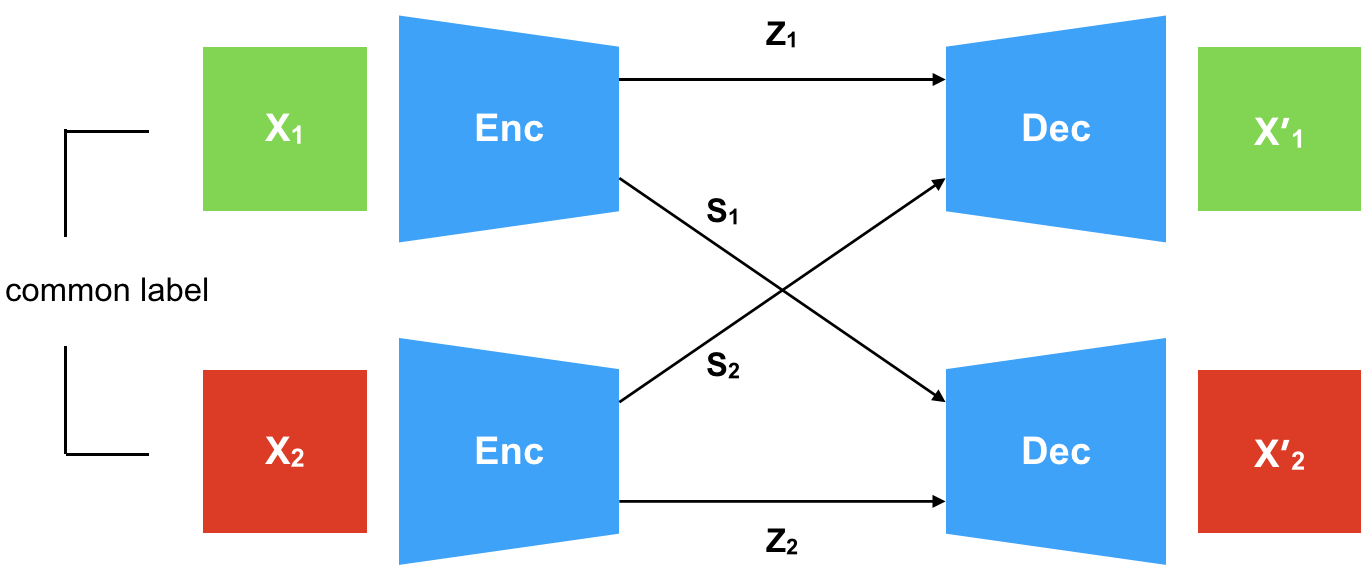}
\caption{Image reconstruction using VAEs by swapping the $s$ latent variable between two images from the same class. This process works with pairwise similarity labels, as we do not need to know the actual class label of the sampled image pair.}
\label{fig:forward_phase}
\end{figure}

\noindent \textbf{Forward cycle.} We sample a pair of images, $x_1$ and $x_2$, from the dataset that have the same class label. We pass both of them through the encoder to generate the corresponding latent representations $Enc(x_1) = (z_1, s_1)$ and $Enc(x_2) = (z_2, s_2)$. The input to the decoder is constructed by swapping the specified latent variables of the two images. This produces the following reconstructions: $x_1' = Dec(z_1, s_2)$ and $x_2' = Dec(z_2, s_1)$. Since both these images share class labels, swapping the specified latent variables should have no effect on the reconstruction loss function. We can re-write the conditional likelihood of the decoder as $p_\theta(x|z,s^*)$, where $s^* = f_s(x^*)$ and $x^*$ is any image with the same class label as $x$. The entire forward cycle minimizes the modified variational upper-bound given in Eq. \ref{eq:4}. Fig. \ref{fig:forward_phase} shows a diagrammatic representation of the forward cycle.

\begin{equation} \label{eq:4}
\underset{Enc, \hspace{0.1cm} Dec}{\min} \hspace{0.1cm}\mathcal{L}_{forward} = - \mathbb{E}_{q_{\phi}(z|x, s^*)}[\text{log} \hspace{0.1cm} p_\theta(x|z,s^*)] + \text{KL}(q_{\phi}(z|x, s^*) \hspace{0.1cm} \| \hspace{0.1cm} p(z))
\end{equation}

It is worth noting that forward cycle does not demand actual class labels at any given time. This results in the requirement of a weaker form of supervision in which images need to be annotated with pairwise similarity labels. This is in contrast with the previous works of Mathieu et al. \cite{DBLP:journals/corr/MathieuZSRL16}, which requires actual class labels, and Szab{\'{o}} et al. \cite{DBLP:journals/corr/abs-1711-02245}, which requires image triplets.

The forward cycle mentioned above is similar to the auto-encoder reconstruction loss presented in \cite{DBLP:journals/corr/MathieuZSRL16} and \cite{DBLP:journals/corr/abs-1711-02245}. As discussed in Sec. \ref{sec:intro}, the forward cycle alone can produce a \emph{degenerate solution} (Fig. \ref{fig:degenerate_ex} (c) and (d)) as there is no constraint which prevents the decoder from reconstructing images using only the unspecified latent variables. In \cite{DBLP:journals/corr/MathieuZSRL16} and \cite{DBLP:journals/corr/abs-1711-02245}, an adversarial loss function has been successfully applied to specifically tackle the \emph{degenerate solution}. The resulting generative model works well, however, adversarial training is challenging in general and has limitations in effectively disentangling the latent space. For now, we defer this discussion to Sec. \ref{sec:adversarial_vs_cycle_consistent}. In the next section, we introduce our non-adversarial method, based on reverse cycle-consistency, to avoid learning a \emph{degenerate solution} and explicitly train the encoder to prevent information associated with specified factors from leaking into the unspecified subspace.

\subsection{Preventing a Degenerate Solution}

\begin{figure}
\centering
\includegraphics[height=3.5cm]{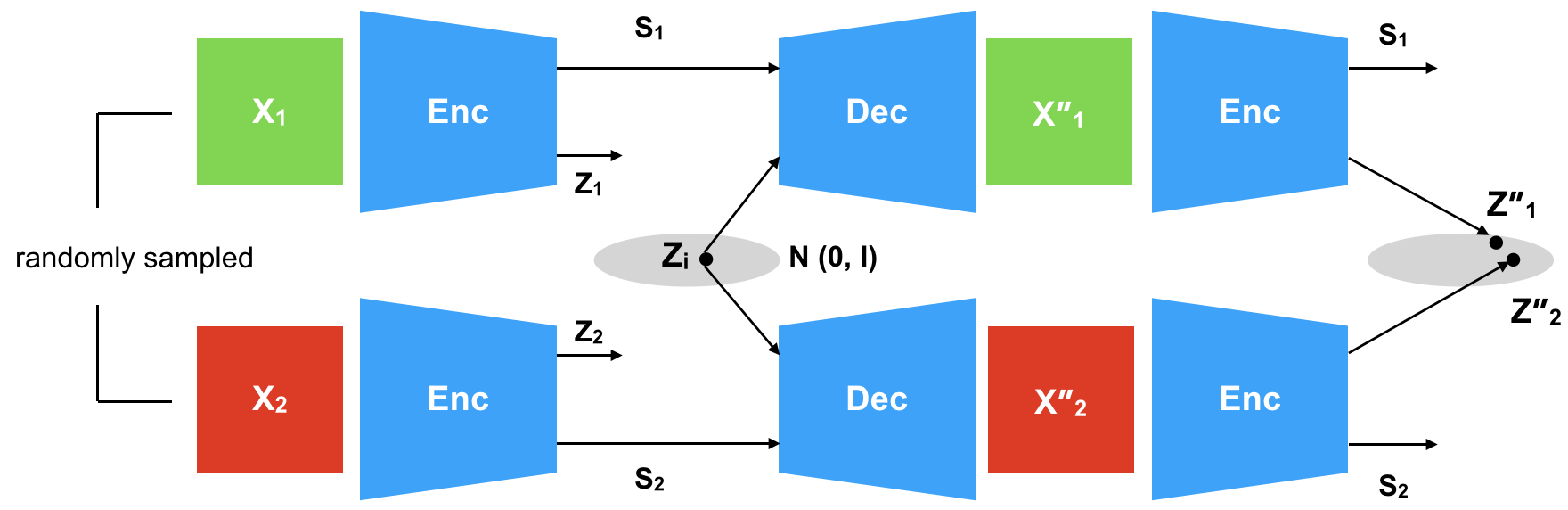}
\caption{Reverse cycle of the cycle-consistent VAE architecture. A point sampled from the $z$ latent space, combined with specified factors from two separate sources, forms two different images. However, we should be able to obtain the same sampled point in the $z$ space if we pass the two generated images back through the encoder.}
\label{fig:backward_phase}
\end{figure}

\noindent \textbf{Reverse cycle.} The reverse cycle shown in Fig. \ref{fig:backward_phase} is based on the idea of cyclic-consistency in the unspecified latent space. We sample a point $z_i$ from the Gaussian prior $p(z)=\mathcal{N}(0,I)$ over the unspecified latent space and pass it through the decoder in combination with specified latent variables $s_1 = f_s(x_1)$ and $s_2 = f_s(x_2)$ to obtain reconstructions $x_1'' = Dec(z_i, s_1)$ and $x_2'' = Dec(z_i, s_2)$ respectively. Unlike the forward cycle, $x_1$ and $x_2$ need not have the same label and can be sampled independently. Since both images $x_1''$ and $x_2''$ are generated using the same $z_i$, their corresponding unspecified latent embeddings $z_1'' = f_z(x_1'')$ and $z_2'' = f_z(x_2'')$ should be mapped close to each other, regardless of their specified factors. Such a constraint promotes marginal independence of $z$ from $s$ as images generated using different specified factors could potentially be mapped to the same point in the unspecified latent subspace. This step directly drives the encoder to produce disentangled representations by only retaining information related to the unspecified factors in the $z$ latent space.

The variational loss in Eq. (\ref{eq:4}) enables sampling of the unspecified latent variables and aids the generation of novel images. However, the encoder does not necessarily learn a unique mapping from the image space to the unspecified latent space. In other words, samples with similar unspecified factors are likely to get mapped to significantly different unspecified latent variables. This observation motivates our \emph{pairwise} reverse cycle loss in Eq. (\ref{eq:5}), which penalizes the encoder if the unspecified latent embeddings $z_1''$ and $z_2''$ have a large pairwise distance, but not if they are mapped farther away from the originally sampled point $z_i$. This modification is in contrast with the typical usage of cycle-consistency in previous works. We found that minimizing the pairwise reverse cycle loss in Eq. (\ref{eq:5}) was easier than its absolute counterpart ($|| z_i - z_1'' || + || z_i - z_2'' ||$), both in terms of the loss value and the extent of disentanglement.

\begin{equation} 
\begin{split}\label{eq:5}
\underset{Enc}{\min} \hspace{0.1cm}\mathcal{L}_{reverse} = \mathbb{E}_{x_1, x_2 \sim p(x), \hspace{0.1cm} z_i \sim \mathcal{N}(0, I)}[|| \hspace{0.1cm} &f_z(Dec(z_i, f_s(x_1))) \\-\hspace{0.1cm} &f_z(Dec(z_i, f_s(x_2))) \hspace{0.1cm} ||_{1}]
\end{split}
\end{equation}

\section{Experiments}\label{sec:experiments}

We evaluate the performance of our model on three datasets: MNIST \cite{Lecun98gradient-basedlearning}, 2D Sprites \cite{DBLP:conf/nips/ReedZZL15,LiberatedPixelCup} and LineMod \cite{Hinterstoisser:2012:MBT:2481913.2481959,DBLP:journals/corr/WohlhartL15}. We divide our experiments into two parts. The first part evaluates the performance of our model in terms of the quality of disentangled representations. The second part evaluates the image generation capabilities of our model. We compare our results with the recent works in \cite{DBLP:journals/corr/MathieuZSRL16,DBLP:journals/corr/abs-1711-02245}. The three dataset we use are described below:

\noindent \textbf{MNIST.} The MNIST dataset \cite{Lecun98gradient-basedlearning} consists of hand-written digits distributed amongst 10 classes. The specified factors in case of MNIST is the digit identity, while the unspecified factors control digit slant, stroke width etc.

\noindent \textbf{2D Sprites.} 2D Sprites consists of game characters (sprites) animated is different poses for use in small scale indie game development. We download the dataset from \cite{LiberatedPixelCup}, which consists of 480 unique characters according to variation in gender, hair type, body type, armor type, arm type and greaves type. Each unique character is associated with 298 different poses, 120 of which have weapons and the remaining do not. In total,  we have 143040 images in the dataset. The training, validation and the test set contain 320, 80 and 80 unique characters respectively. This implies that character identity in each of the training, validation and test split is mutually exclusive and the dataset presents an opportunity to test our model on completely unseen object identities. The specified factors latent space for 2D Sprites is associated with the character identity, while the pose is associated with the unspecified factors.

\noindent \textbf{Line-MOD.} LineMod \cite{Hinterstoisser:2012:MBT:2481913.2481959} is an object recognition and 3D pose estimation dataset with 15 unique objects:  `ape', `benchviseblue', `bowl', `cam', `can', `cat', `cup', `driller', `duck', `eggbox', `glue', `holepuncher', `iron', `lamp' and `phone', photographed in a highly cluttered environment. We use the synthetic version of the dataset \cite{DBLP:journals/corr/WohlhartL15}, which has the same objects rendered under different viewpoints. There are 1541 images per category and we use a split of 1000 images for training, 241 for validation and 300 for test. The specified factors latent space models the object identity in this dataset. The unspecified factors latent space models the remaining factors of variation in the dataset.

During forward cycle, we randomly pick image pairs defined by the same specified factors of variation. During reverse cycle, the selection of images is completely random. All our models were implemented using PyTorch \cite{paszke2017automatic}. We include the specific details about our architectures in the supplementary material section. 

\subsection{Quality of Disentangled Representations}\label{sec:adversarial_vs_cycle_consistent}

\begin{table}[htb]
\fontsize{8}{9}\selectfont
\begin{center}
\begin{tabular}{ |p{2.2cm}|c|c|c|c|c|c| } 
\hline
\textbf{Architecture} & \textbf{$z$ dim} & \textbf{$s$ dim} & \textbf{$z$ train acc.} & \textbf{$z$ test acc.} & \textbf{$s$ train acc.} & \textbf{$s$ test acc.} \\
\hline
\multicolumn{7}{|c|}{MNIST} \\
\hline
Szab{\'{o}} et al. & 16 & 16 & 97.65 & 96.08 & 98.89 & 98.46 \\
Mathieu et al. & 16 & 16 & 70.85 & 66.83 & 99.37 & 98.52 \\ 
Ours & 16 & 16 & \textbf{17.72} & \textbf{17.56} & 99.72 & 98.35 \\ 
\hline
Szab{\'{o}} et al. & 64 & 64 & 99.69 & 98.14 & 99.41 & 98.05 \\ 
Mathieu et al. & 64 & 64 & 74.94 & 72.20 & 99.94 & 98.64 \\
Ours & 64 & 64 & \textbf{26.04} & \textbf{26.55} & 99.95 & 98.33 \\ \hline
\multicolumn{7}{|c|}{2D Sprites} \\
\hline
Szab{\'{o}} et al. & 512 & 64 & 99.72 & 99.63 & 99.85 & 99.79 \\
Mathieu et al. & 512 & 64 & 12.05 & 11.98 & 99.18 & 96.75 \\ 
Ours & 512 & 64 & 11.55 & 11.47 & 98.53 & 97.16 \\ 
\hline
Szab{\'{o}} et al. & 1024 & 512 & 99.79 & 99.65 & 99.87 & 99.76 \\ 
Mathieu et al. & 1024 & 512 & 12.48 & 12.25 & 99.22 & 97.45 \\
Ours & 1024 & 512 & 11.27 & 11.61 & 98.13 & 97.22 \\ \hline
\multicolumn{7}{|c|}{LineMod} \\
\hline
Szab{\'{o}} et al. & 64 & 256 & 100.0 & 100.0 & 100.0 & 100.0 \\
Mathieu et al. & 64 & 256 & 90.14 & 89.17 & 100.0 & 100.0 \\ 
Ours & 64 & 256 & \textbf{62.11} & \textbf{57.17} & 99.99 & 99.86 \\ 
\hline
Szab{\'{o}} et al. & 256 & 512 & 100.0 & 99.97 & 100.0 & 100.0 \\ 
Mathieu et al. & 256 & 512 & 86.87 & 86.46 & 100.0 & 100.0 \\
Ours & 256 & 512 & \textbf{60.34} & \textbf{57.70} & 100.0 & 100.0 \\ \hline
\end{tabular}

\end{center}
\caption{Quantitative results for the three datasets. Classification accuracies on the $z$ and $s$ latent spaces are a good indicator of the amount of specified factor information present in them. Since we are aiming for disentangled representations for unspecified and specified factors of variation, \emph{lower is better} for the $z$ latent space and \emph{higher is better} the $s$ latent space.}
\label{table:1}
\end{table}

\begin{figure}
\centering
\includegraphics[height=3.0cm]{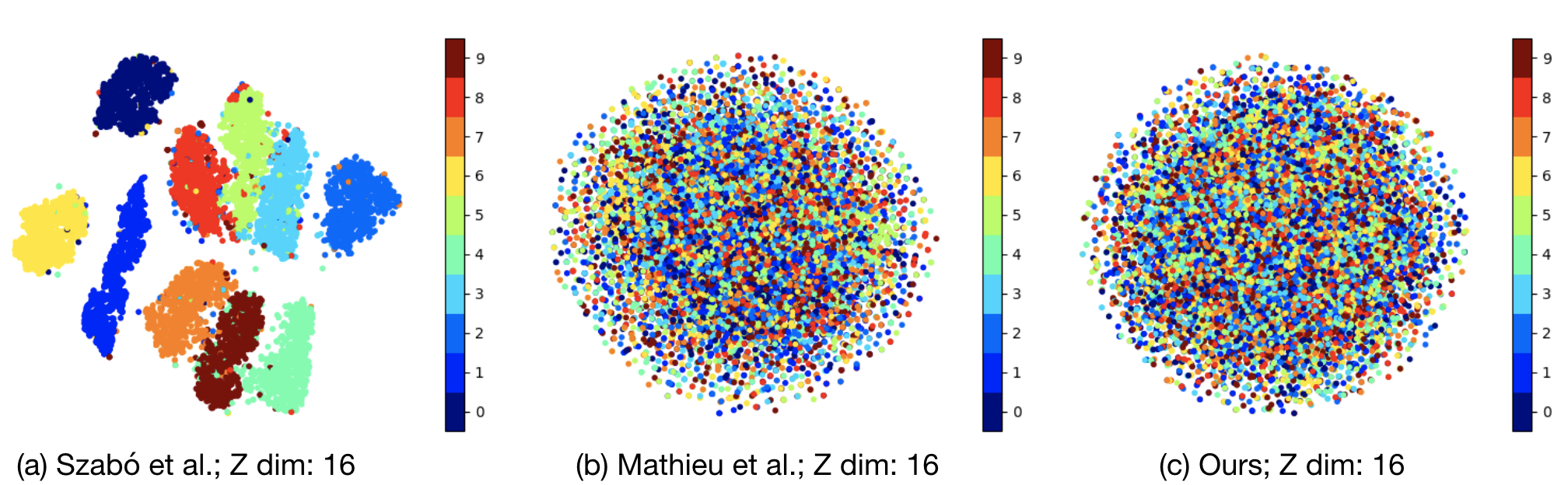}
\caption{Comparison between t-SNE plots of the $z$ latent space for MNIST. We can see good cluster formation according to class identities in (a) \cite{DBLP:journals/corr/abs-1711-02245}, indicating that adversarial training alone does not promote marginal independence of $z$ from $s$. Mathieu's work \cite{DBLP:journals/corr/MathieuZSRL16} in (b) uses re-parameterization on the encoder output to create confusion regarding the specified factors in the $z$ space while retaining information related to the unspecified factors. Our work (c) combines re-parameterization with reverse cycle loss to create confusion regarding the specified factors.}
\label{fig:t_sne}
\end{figure}

We set up the quantitative evaluation experiments similar to \cite{DBLP:journals/corr/MathieuZSRL16}. We train a two layer neural network classifier separately on the specified and unspecified latent embeddings generated by each competing model. Since the specified factors of variation are associated with the available labels in each dataset, the classifier accuracy gives a fair measure of the information related to specified factors of variation present in the two latent subspaces. If the factors were completely disentangled, we expect the classification accuracy in the specified latent space to be perfect, while that in the unspecified latent space to be close to chance. In this experiment, we also investigate the effect of change in the dimensionality of the latent spaces. We report the quantitative comparisons in Table \ref{table:1}.

The quantitative results in Table \ref{table:1} show consistent trends for our proposed Cycle-Consistent VAE architecture across all the three datasets as well as for different dimensionality of the latent spaces. Classification accuracy in the unspecified latent subspace is the smallest for the proposed architecture, while it is comparable with the others in the specified latent subspace. These trends indicate that among the three competing models, the proposed one leaks the least amount of specified factor information into the unspecified latent subspace. This restricted amount of leakage of specified information can be attributed to the reverse cycle-consistency loss that explicitly trains the encoder to disentangle factors more effectively.

We also visualize the unspecified latent space as t-SNE plots \cite{Maaten08visualizingdata} to check for the presence of any apparent structure based on the available labels with the MNIST dataset. Fig. \ref{fig:t_sne} shows the t-SNE plots of the unspecified latent space obtained by each of the competing models. The points are color-coded to indicate specified factor labels, which in case of MNIST are the digit identities. We can see clear cluster structures in Fig. \ref{fig:t_sne} (a) indicating strong presence of the specified factor information in the unspecified latent space. This observation is consistent with the quantitative results shown in Table \ref{table:1}. As shown in Fig. \ref{fig:t_sne} (b) and (c), the t-SNE plots for Mathieu et al.'s model \cite{DBLP:journals/corr/MathieuZSRL16} and our model appear to have similar levels of confusion with respect to the specified factor information. However, since t-SNE plots are approximations, the quantitative results reported in Table. \ref{table:1} better capture the performance comparison.

The architectures in \cite{DBLP:journals/corr/MathieuZSRL16,DBLP:journals/corr/abs-1711-02245} utilize adversarial training in combination with a regular and a variational auto-encoder respectively. Despite the significant presence of specified factor information in the unspecified latent embeddings from Szab{\'{o}} et al.'s model \cite{DBLP:journals/corr/abs-1711-02245}, it successfully generates novel images by combining the specified and unspecified factors (shown in Sec. \ref{sec:image_generation}). This apparently conflicting observation suggests that the decoder somehow learns to ignore the specified factor information in the unspecified latent space. We conjecture that since the adversarial loss updates the decoder and the encoder parameters together, and in that order, the encoder remains less likely to disentangle the latent spaces.

A similar argument can be made that Mathieu et al.'s \cite{DBLP:journals/corr/MathieuZSRL16} architecture does not explicitly train the encoder to disentangle factors of variation, thus resulting in higher classification accuracy in the unspecified latent space. This behavior, however, is mitigated to a large extent due to the VAE framework, which promotes class confusion in the unspecified latent subspace by performing reparametrization at the time of new image generation. Our approach benefits from the reparametrization as well, however, significantly lower classification accuracies on the unspecified latent space embeddings indicate that the encoder learns to disentangle the factors better by minimizing the reverse cycle-consistency loss.

\subsection{Quality of Image Generation}\label{sec:image_generation}

\begin{figure}
\centering
\includegraphics[height=6.0cm,width=8.4cm]{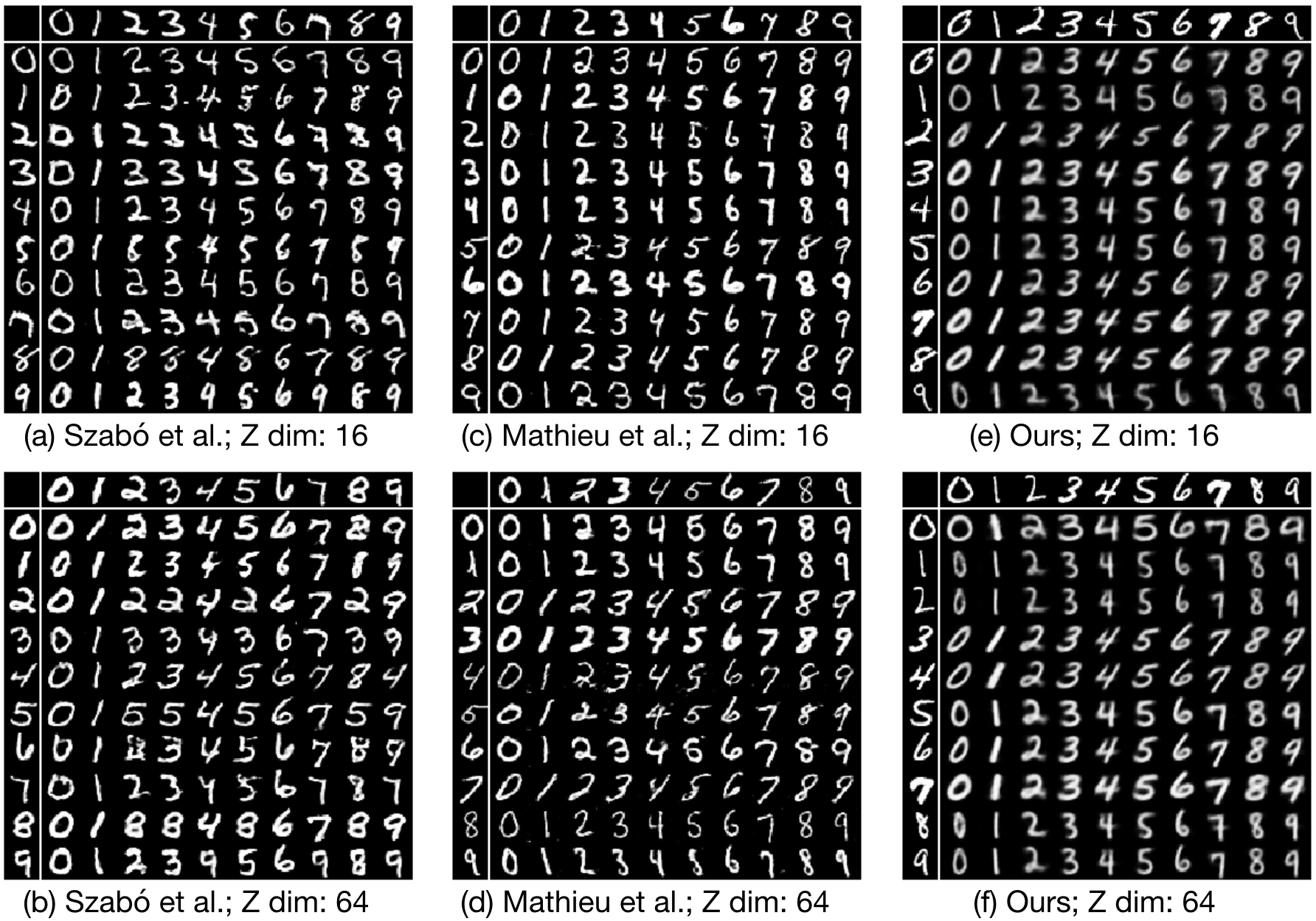}
\caption{Image generation results on MNIST by swapping $z$ and $s$ variables. The top row and the first column are randomly selected from the test set. The remaining grid is generated by taking $z$ from the digit in first column and $s$ from the digit in first row. This keeps the unspecified factors constant in rows and the specified factors constant in columns.}
\label{fig:mnist_swap}
\end{figure}

The quality of image generation is evaluated in three different setups. First, we test the capability of our model to combine unspecified and specified latent variables from different sources or images to generate a new image. This experiment is done in form of a grid of images, where the first row and the first column is taken from the test set. The remaining grid is filled up with image generated by combining the specified factor of variation from images in the first row and the unspecified factors of variation from images in the first column. For this evaluation, we compare our results against the images generated by prior works \cite{DBLP:journals/corr/MathieuZSRL16} and \cite{DBLP:journals/corr/abs-1711-02245}. Unlike the non-adversarial approach proposed by Szab{\'{o}} et al. \cite{DBLP:journals/corr/abs-1711-02245}, our model is robust to the choices of dimensionality for both $z$ and $s$ variables. Hence, we show that our model avoids \emph{degeneracy} for significantly higher dimensions of latent variables, in comparison to the base values, despite being a non-adversarial architecture. Second, we show the variation captured in the two latent manifolds of our models by linear interpolation. The images in the top-left and the bottom-right corner are taken from the test set and similar to the first evaluation, the remaining images are generated by keeping $z$ constant across the rows and $s$ constant across the columns. And lastly, we check the conditional image generation capability of our model by conditioning on the $s$ variable and sampling data points directly from the Gaussian prior $p(z)$ for the $z$ variable.

\begin{figure}
\centering
\includegraphics[height=8.0cm]{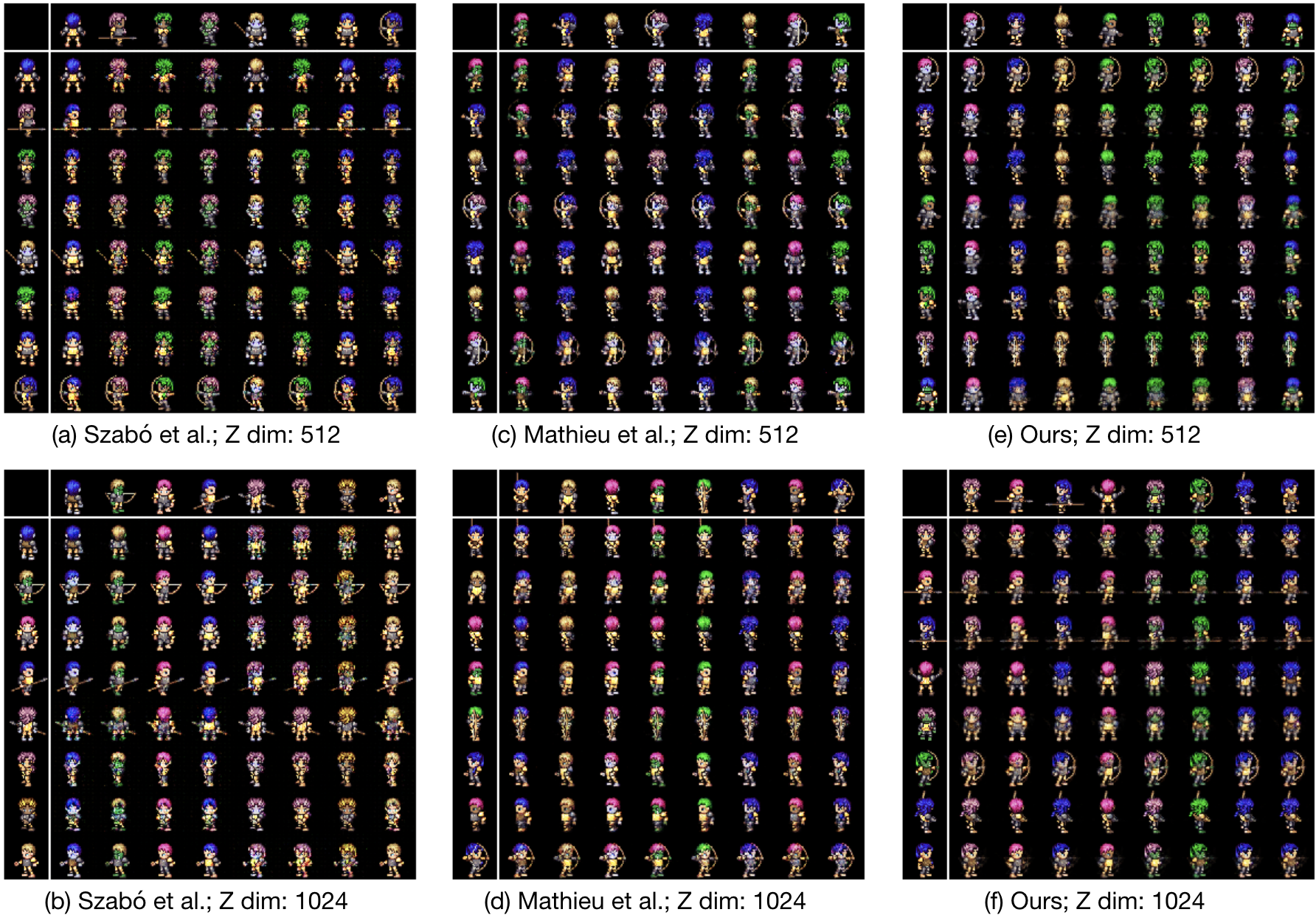}
\caption{Image generation results on 2D Sprites by swapping $z$ and $s$ variables. Arrangement of the grid is same as Fig. \ref{fig:mnist_swap}.}
\label{fig:sprites_swap_1}
\end{figure}

\begin{figure}
\centering
\includegraphics[height=8.0cm,width=11.2cm]{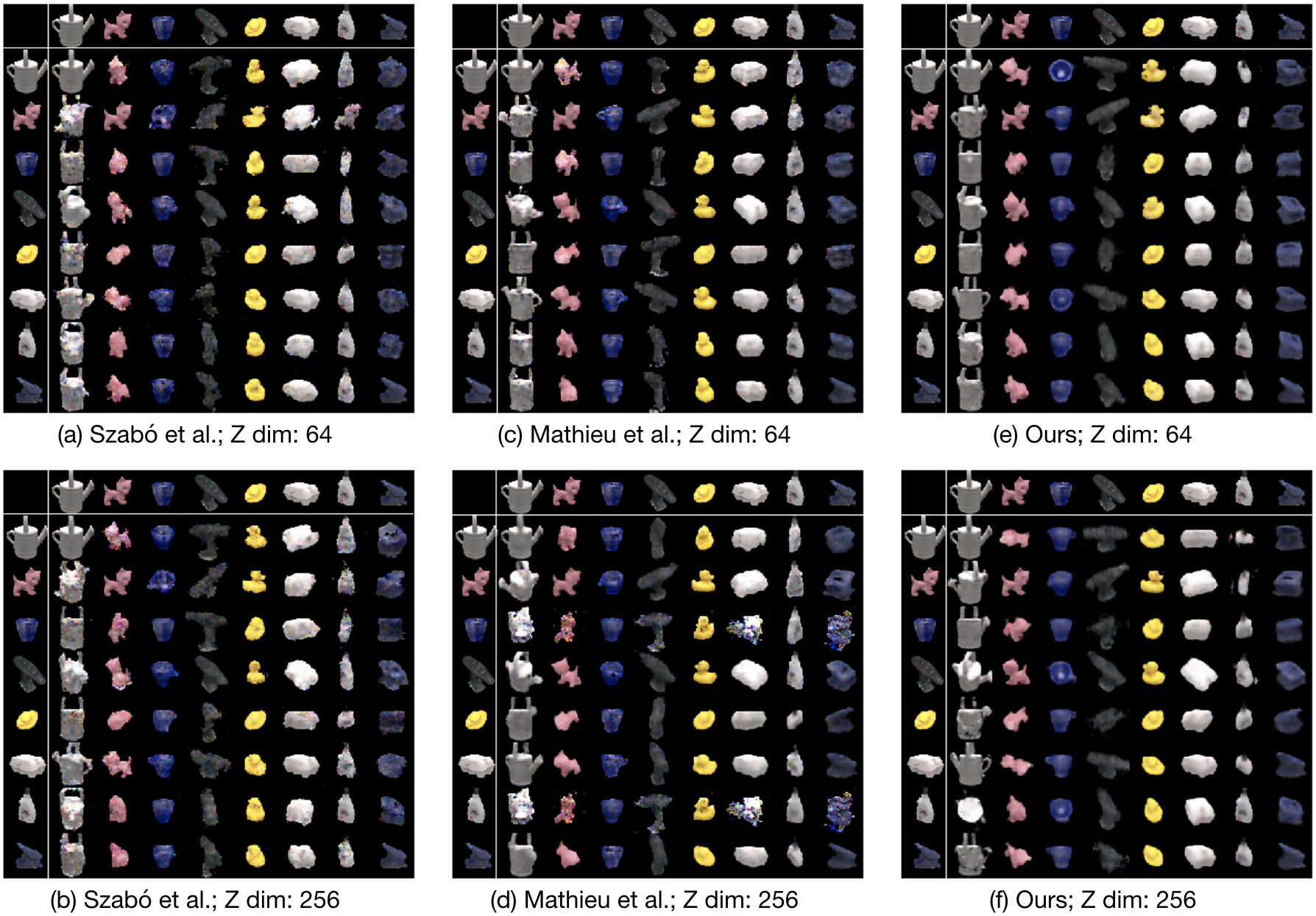}
\caption{Image generation results on LineMod by swapping $z$ and $s$ variables. Arrangement of the grid is same as Fig. \ref{fig:mnist_swap}. As explained in Sec. \ref{sec:image_generation}, we do not observe a direct transfer of viewpoint between the objects.}
\label{fig:linemod_swap}
\end{figure}

The first evaluation of generating new images by combining $z$ and $s$ from different sources is shown in Figures \ref{fig:mnist_swap}, \ref{fig:sprites_swap_1} and \ref{fig:linemod_swap}. LineMod dataset does not have a fixed alignment of objects for the same viewpoint. For example, an image of a `duck' will not be aligned in the same direction as an image of a `cat' for a common viewpoint. Also, our assumption that viewpoint is the only factor of variation associated with the unspecified space does not hold true for LineMod due to the complex geometric structure of each object. Hence, as is apparent from Fig. \ref{fig:linemod_swap}, interpretation of transfer of unspecified factors as viewpoint transfer does not exactly hold true. For a direct comparison of the transfer of unspecified factors between different models, we keep the test images constant across the different image grids shown for LineMod.

\begin{figure}
\centering
\includegraphics[height=3.5cm]{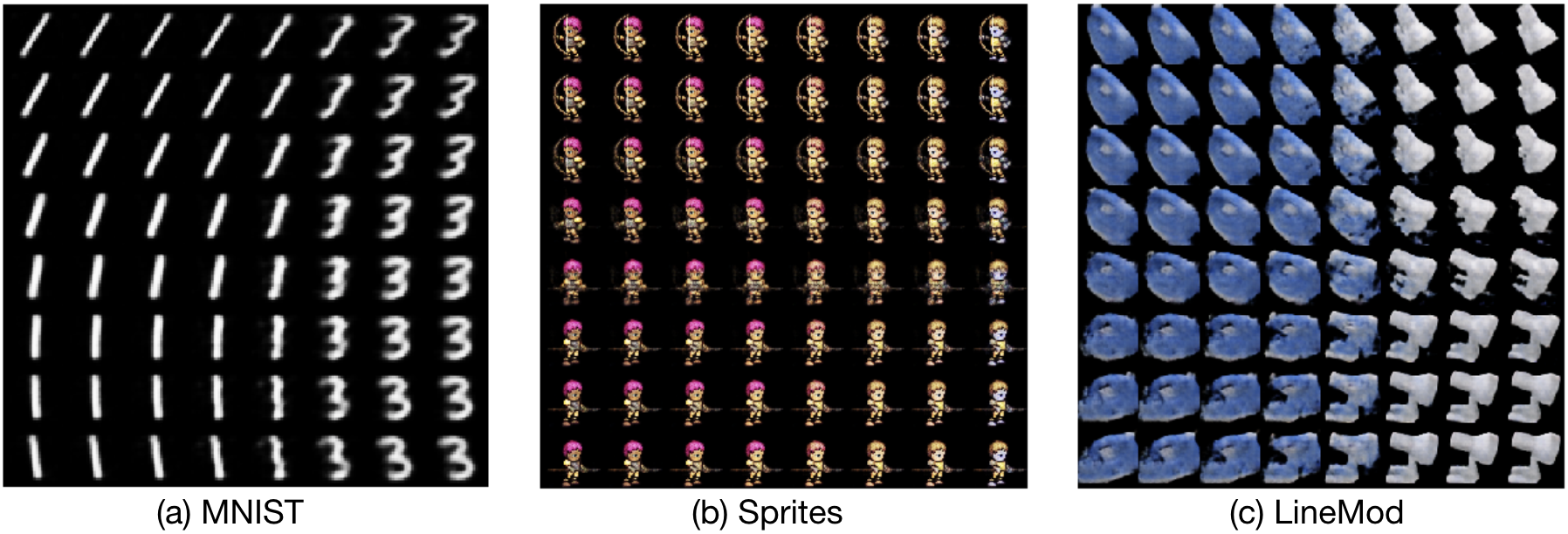}
\caption{Linear interpolation results for our model in the $z$ and $s$ latent spaces. The images in the top-left and the bottom-right corner are taken from the test set. Like Fig. \ref{fig:mnist_swap}, $z$ variable is constant in the rows, while $s$ is constant in the columns.}
\label{fig:linear_interpolation}
\end{figure}

\begin{figure}
\centering
\includegraphics[height=2.5cm]{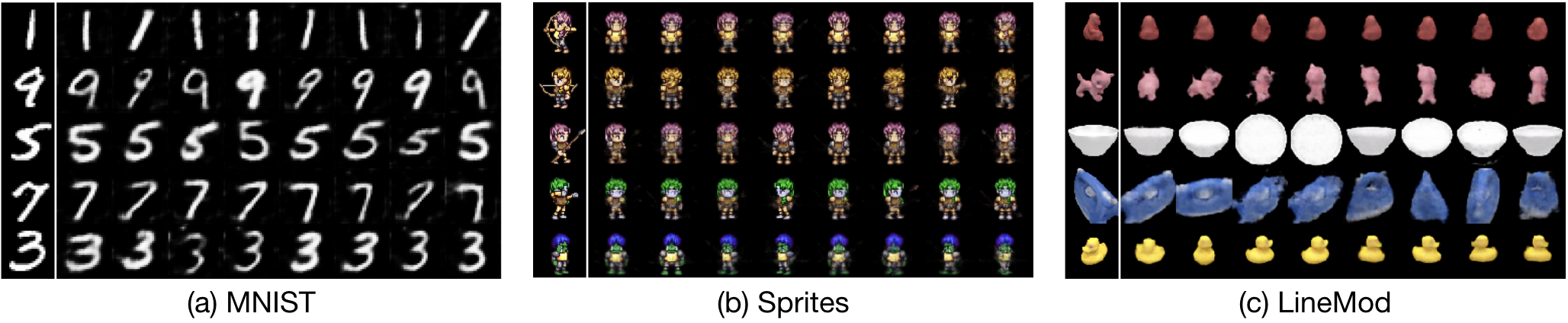}
\caption{Image generation by conditioning on $s$ variable, taken from test images, and sampling the $z$ variable from $\mathcal{N}(0, I)$.}
\label{fig:sampling}
\end{figure}

Fig. \ref{fig:linear_interpolation} shows the result of linear interpolation of the latent manifolds learned by our model for three datasets. Fig. \ref{fig:sampling} shows the result of conditional image generation by sampling directly from the prior $p(z)$.

\section{Conclusion}\label{sec:conclusion}

In this paper we introduced a simple yet effective way to disentangle specified and unspecified factors of variation by leveraging the idea of cycle-consistency. The proposed architecture needs only weak supervision in the form of pairs of data having similar specified factors. The architecture does not produce degenerate solutions and is robust to the choices of dimensionality of the latent space. Through our experimental evaluations, we found that even though adversarial training produces good visual reconstructions, the encoder does not necessarily learn to disentangle the factors of variation effectively. Our model, on the other hand, achieves compelling quantitative results on three different datasets and shows good image generation capabilities as a generative model.

We also note that generative models based on VAEs produce less sharper images compared to GANs and our model is no exception. One way to address this problem could be to train our cycle-consistent VAE as the first step, followed by training the decoder with a combination of adversarial and reverse cycle-consistency loss. This training strategy may improve the sharpness of the generated images while maintaining the disentangling capability of the encoder. Another interesting direction to pursue from here would be to further explore the methods that disentangle factors of variation without using any form of supervision.

\bibliographystyle{splncs}

\title{Supplementary Material}

\author{}
\institute{}

\maketitle

\section*{Algorithm}

The following algorithm summarizes the entire training procedure. The notations used here have been introduced in Sec. 3 of the main paper, while the loss functions are from Eq. \ref{eq:4} and \ref{eq:5}.

\begin{table}[h!]
\centering
 \begin{tabular}{l c} 
 \hline
 \textbf{Algorithm 1} \\ [0.5ex] 
 \hline\hline
 for i in 1...n training iterations \\ 
 \hspace{0.5cm} \textbf{Train forward cycle} \\
 \hspace{0.5cm} Sample an image pair ($x_1$, $x_2$) according to pairwise similarity labels \\
 \hspace{0.5cm} Compute latent embeddings ($\mu_1, \sigma_1, s_1) = Enc(x_1)$ and $(\mu_2, \sigma_2, s_2) = Enc(x_2)$\\
 \hspace{0.5cm} Sample $z_1 \sim \mathcal{N}(\mu_1,\sigma_1)$ and $z_2 \sim \mathcal{N}(\mu_2,\sigma_2)$ \\
 \hspace{0.5cm} Compute reconstructions $x_1' = Dec(z_1, x_2)$ and $x_2' = Dec(z_2, x_1)$\\
 \hspace{0.5cm} Compute KL-divergence loss for ($\mu_1, \sigma_1
 $) and ($\mu_2, \sigma_2$) independently \\
 \hspace{0.5cm} Compute L2 reconstruction loss between ($x_1'$ and $x_1$) and ($x_2'$ and $x_2$) \\
 \hspace{0.5cm} Back-propagate the gradients to train both Enc and Dec \\ [1ex]
 \hspace{0.5cm} \textbf{Train reverse cycle} \\ 
 \hspace{0.5cm} Sample any two images $x_1$ and $x_2$ from the dataset \\  
 \hspace{0.5cm} Compute specified factors latent embeddings $s_1 = f_s(x_1)$ and $s_2 = f_s(x_2)$ \\
 \hspace{0.5cm} Sample $z_i \sim \mathcal{N}(0, I)$ \\
 \hspace{0.5cm} Compute reconstructions $x_1'' = Dec(z_i, s_1)$ and $x_2'' = Dec(z_i, s_2)$ \\
 \hspace{0.5cm} Compute unspecified factors latent embeddings $(\mu_1'',\sigma_1'') = f_z(x_1'')$ \\\hspace{1.0cm} and $(\mu_2'',\sigma_2'') = f_z(x_2'')$ \\
 \hspace{0.5cm} Assign the computed means to $z_1'' = \mu_1''$ and $z_2'' = \mu''$ \\
 \hspace{0.5cm} Compute L1 reconstruction loss between $z_1''$ and $z_2''$ \\
 \hspace{0.5cm} Back-propagate the gradients to train only Enc \\ [1ex]
 \hline
 \end{tabular}
\label{table:2}
\end{table}

\section*{Network Architectures}

The Encoder consists of a common convolutional trunk that splits into two branches of fully-connected nodes in the last layer in order to output latent embeddings for the specified and unspecified factors of variation. The fully-connected nodes of the unspecified latent space are further split in two parts, as they output both mean and variance to parameterize the approximate posterior. The common convolutional trunk consists of \emph{conv blocks}, each with a convolutional, instance normalization and ReLU layers. Instead of using \emph{max-pooling} to reduce the spatial dimensions of feature maps, we use convolutional layers with a stride of 2.

The Decoder contains two branches of fully-connected nodes in the initial layer, which take inputs from the corresponding $z$ and $s$ latent embeddings. These are then concatenated together and reshaped in order to be passed through a series of \emph{conv blocks}, each of which consist of convolutional, instance normalization and ReLU layers again. However, unlike the Encoder, convolutional layers in the Decoder have partial strides to perform upsampling in the spatial dimensions of the feature maps.

The initial dimensions of an image from the MNIST dataset is 28x28x1. In the Encoder, we use 3 \emph{conv blocks} each containing convolutional layer with a filter size of 5 and stride 2. Similarly, the Decoder uses 3 \emph{conv blocks} to take latent embeddings back to the size of the original image. An image from either 2D Sprites or LineMod is of size 64x64x3. For these, we use 4 \emph{conv blocks} in the same filter size and stride configuration, for both Encoder and Decoder.

\section*{t-SNE Plots}

Here, we show visualizations of the unspecified latent space as t-SNE plots \cite{Maaten08visualizingdata} for 2D Sprites \cite{DBLP:conf/nips/ReedZZL15,LiberatedPixelCup} and LineMod datasets \cite{Hinterstoisser:2012:MBT:2481913.2481959,DBLP:journals/corr/WohlhartL15}. The points are color-coded according to their specified factors label. 

\begin{figure}
\centering
\includegraphics[height=3.0cm]{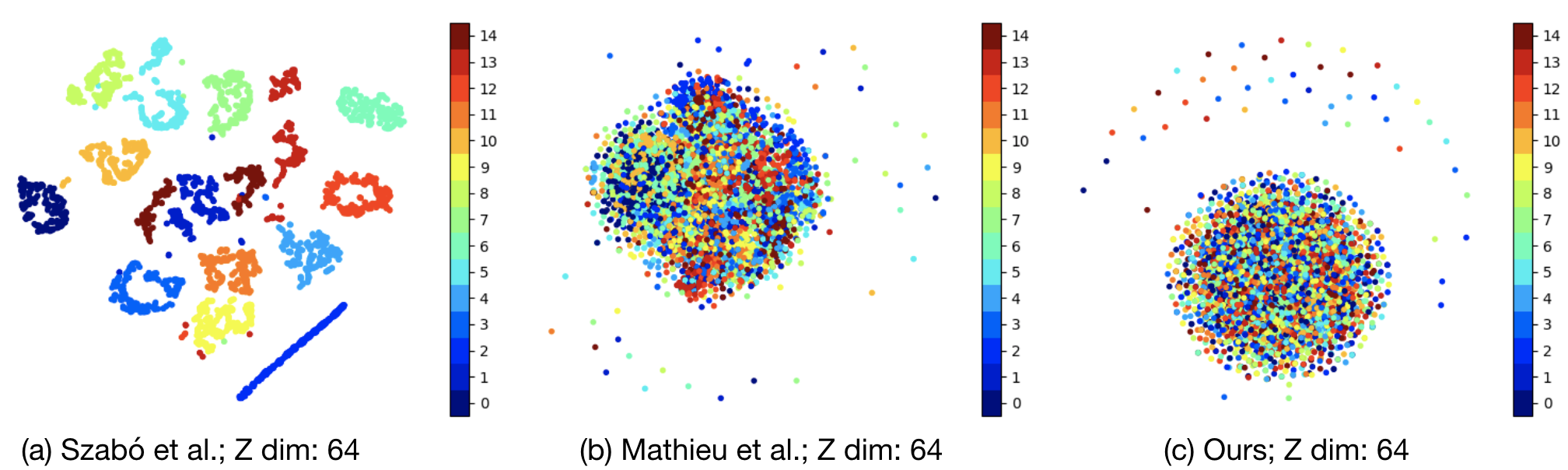}
\caption{Comparison between t-SNE plots of the $z$ latent space for LineMod. We can see good cluster formation according to class identities in (a) \cite{DBLP:journals/corr/abs-1711-02245}, indicating that adversarial training alone does not promote marginal independence of $z$ from $s$. Mathieu's work \cite{DBLP:journals/corr/MathieuZSRL16} in (b) uses re-parameterization on the encoder output to create confusion regarding the specified factors in the $z$ space while retaining information related to the unspecified factors. Our work (c) combines re-parameterization with reverse cycle loss to create confusion regarding the specified factors.}
\label{fig:t_sne_linemod}
\end{figure}

Similar to MNIST (Fig. \ref{fig:t_sne} in the main paper), we observe cluster formation in Fig. \ref{fig:t_sne_linemod} (a) according to the specified factor labels, thus indicating the presence of specified factor information in the unspecified factors space. Observations in Fig. \ref{fig:t_sne_linemod} (b) and (c) have class confusion in the unspecified factors space as expected, an explanation for which has been provided in Sec. \ref{sec:adversarial_vs_cycle_consistent} of the main paper.

\begin{figure}
\centering
\includegraphics[height=3.0cm]{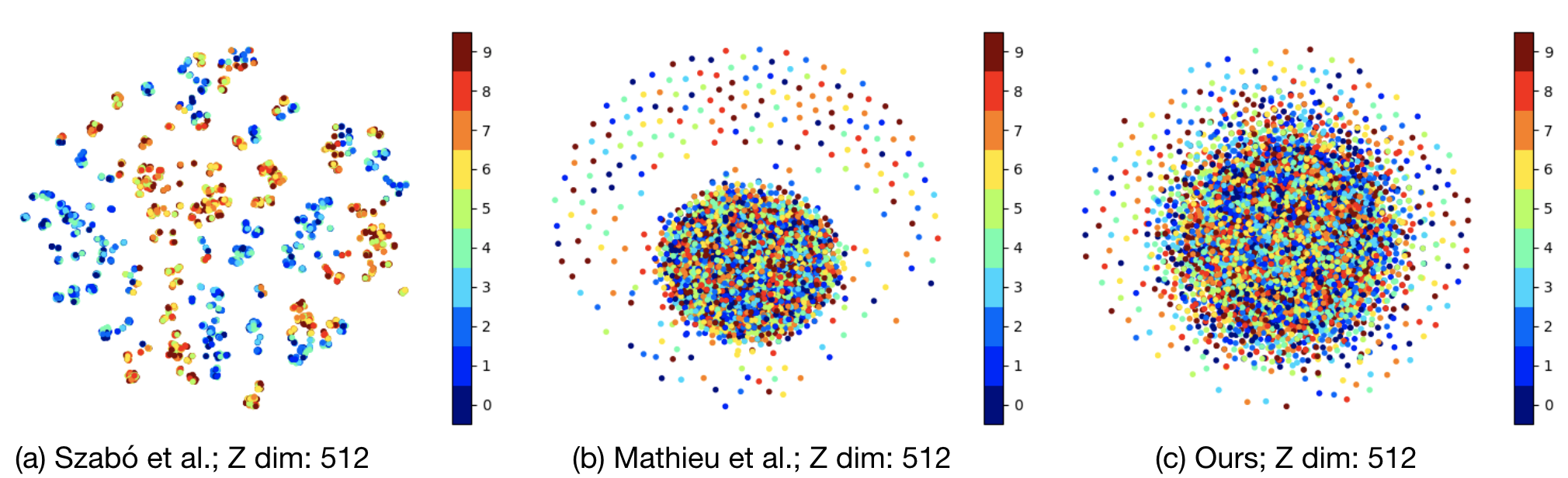}
\caption{Comparison between t-SNE plots of the $z$ latent space for 2D Sprites.}
\label{fig:t_sne_sprites}
\end{figure}

2D Sprites contains 480 unique characters in total, which are categorized into 10 broad classes based on gender and body type of each character. The plot in Fig. \ref{fig:t_sne_sprites} (a) is specifically interesting to us, as even without any clear cluster formation, high classification accuracies in the unspecified latent space for \cite{DBLP:journals/corr/abs-1711-02245} indicate that the classes are clearly separable. The sparseness of the plot alludes to this contrasting observation.

\end{document}